\documentclass{article} 
\usepackage{iclr2026_conference,times}

\usepackage{amsmath,amsfonts,bm}

\def\eqref#1{equation~\ref{#1}}

\def\1{\bm{1}}

\DeclareMathAlphabet{\mathsfit}{\encodingdefault}{\sfdefault}{m}{sl}
\SetMathAlphabet{\mathsfit}{bold}{\encodingdefault}{\sfdefault}{bx}{n}

\usepackage{url}

\usepackage{xcolor}         
\definecolor{sbblue}{HTML}{4878d0}
\definecolor{sbred}{HTML}{d65f5f}
\definecolor{sbgreen}{HTML}{6acc64}
\definecolor{sbbluedeep}{HTML}{4c72b0}
\definecolor{sbreddeep}{HTML}{c44e52}
\definecolor{sbgreendeep}{HTML}{55a868}

\usepackage[T1]{fontenc}

\usepackage{microtype}
\usepackage{amsmath,amssymb,mathtools}
\usepackage{hyperref}
\hypersetup{
  colorlinks,
  citecolor=sbblue,
  linkcolor=sbred,
  urlcolor=sbblue}
\usepackage{cleveref}
\usepackage{tabularx}  
\usepackage{caption}   
\usepackage{booktabs,tabularx,siunitx}
\usepackage[utf8]{inputenc}
\usepackage{float}

\usepackage[hang,flushmargin]{footmisc} 
\DeclareUnicodeCharacter{2212}{-}   
\DeclareUnicodeCharacter{2009}{\,}  
\DeclareUnicodeCharacter{00A0}{~}   

\title{MATH-Beyond: A Benchmark for RL to Expand Beyond the Base Model}

\author{%
    Prasanna Mayilvahanan$^{1,2,3}$ \quad Ricardo Dominguez-Olmedo$^{2,3}$ \\ \\
    \textbf{Thaddäus Wiedemer}$^{1,2,3}$ \quad \textbf{Wieland Brendel}$^{2,3,4}$
}

\iclrfinalcopy 
\begin{document}
\enlargethispage{2\baselineskip} 

\maketitle

{\renewcommand*{\thefootnote}{}\footnotetext{\footnotesize
$^1$University of Tübingen, $^2$Tübingen AI Center, $^3$Max-Planck-Institute for Intelligent Systems, Tübingen, $^4$ELLIS Institute Tübingen. 
Contact: prasanna.mayilvahanan@uni-tuebingen.de. 
Code available at \\ \url{https://brendel-group.github.io/math-beyond/}.}}
\addtocounter{footnote}{-1} 

\begin{abstract}
With the advent of DeepSeek-R1, a new wave of reinforcement learning (RL) methods has emerged that seem to unlock stronger mathematical reasoning. However, a closer look at the open-source ecosystem reveals a critical limitation: with sufficiently many draws (e.g., $\texttt{pass@1024}$), many existing base models already solve nearly all questions on widely used math benchmarks such as MATH-500 and AIME 2024. This suggests that the RL fine-tuning methods prevalent in the LLM reasoning literature largely sharpen existing solution modes rather than discovering entirely new ones. Such sharpening stands in contrast to the broader promise of RL: to foster exploration and to acquire new skills. To move beyond this plateau, we introduce MATH-Beyond (MATH-B), a benchmark deliberately constructed to defeat common open-source models of up to 8B parameters even under large sampling budgets. Improving performance on our benchmark via RL requires methods that learn to reason in ways that go beyond base model capabilities in repeated sampling. Since the problems are drawn from subsets of DAPO-Math-17K and DeepScaleR datasets, they remain topically equivalent to standard high-school math. Validating our premise, RL fine-tuned models such as Nemotron-Research-Reasoning-Qwen-1.5B and DeepScaleR-1.5B-Preview perform poorly on MATH-B at $\texttt{pass@1024}$, showing how existing approaches fall short on tackling harder instances. We hope MATH-B will catalyze exploration-driven RL approaches that elicit deeper reasoning capabilities. We release MATH-B at \url{https://huggingface.co/datasets/brendel-group/MATH-Beyond}.
\end{abstract}

\begin{figure}[H]
    \centering
    \includegraphics[width=1.0
    \linewidth]{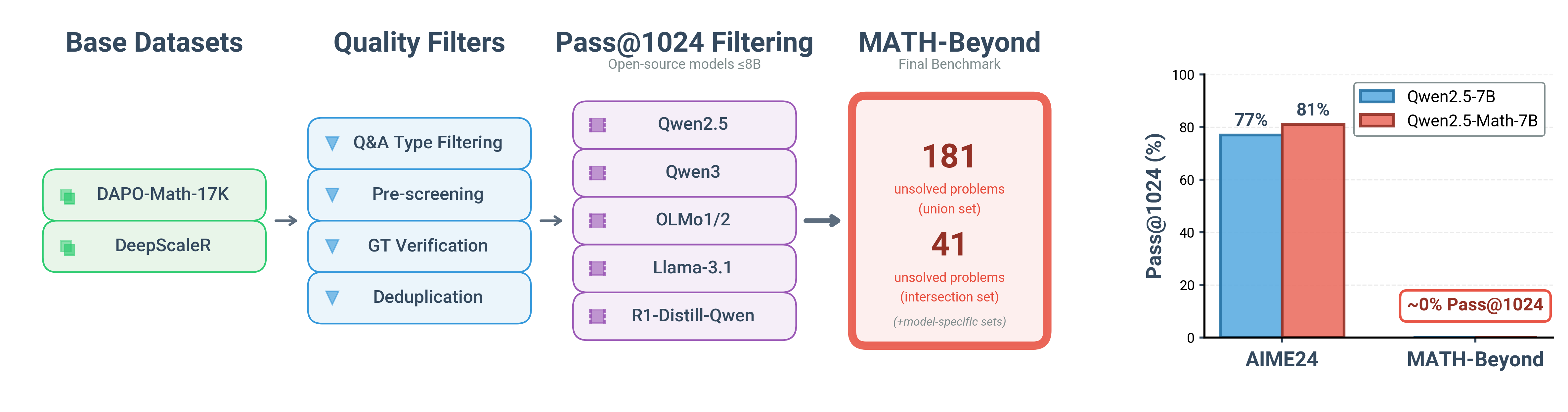}
    \caption{\textbf{MATH-Beyond: Benchmark Construction and Difficulty.} \textit{Left:} Schematic of the MATH-B creation process. A large set of problems from DAPO-Math-17K and DeepScaleR is first refined through quality filters to ensure answer correctness and verifiability. This is followed by evaluation against a gauntlet of open-source base models~($\leq$ 8B, e.g., Qwen3, Qwen2.5 (-Math), DeepSeek-R1-Distill) at a \texttt{pass@1024} budget to isolate problems that lie beyond their limits. The filtering yields the \textsc{MATH-B} suite of benchmarks: a 41-problem intersection set (unsolved by all base models) for evaluating universal difficulty, and a larger 181-problem union set (unsolved by at least one model) with model-specific splits for targeted analysis. This suite provides a rigorous testbed to drive the development of exploration methods for RL. \textit{Right:} An illustration of MATH-B's significant difficulty compared to common test sets like AIME24. Representative open-source models like Qwen2.5 achieve near-zero \texttt{pass@1024} scores on MATH-B, highlighting its difficulty. Qwen2.5 results are from \cite{yue2025doesreinforcementlearningreally}.}
    \label{fig:schematic}
\end{figure}

\section{Introduction}
In the 2010s, deep reinforcement learning showcased its power through striking demonstrations of exploration and skill acquisition~\citep{mnih2013playingatarideepreinforcement}. Atari agents, starting from random play, mastered complex games by discovering strategies unreachable to base policies, guided by exploration incentives and intrinsic rewards~\citep{Ladosz_2022,amin2021survey}. Methods such as count-based exploration~\citep{bellemare2016unifying} and later Go-Explore~\citep{ecoffet2021goexplorenewapproachhardexploration} drove dramatic jumps from inept play to expertise, highlighting RL’s ability to uncover new capabilities. Around the same time, AlphaGo~\citep{Gogame} and AlphaZero~\citep{silver2017masteringchessshogiselfplay} extended this promise to board games like Go, Chess, and Shogi, where self-play took agents from scratch to superhuman mastery, revealing novel strategies along the way.

Against this backdrop, academic progress in reasoning-focused LLMs has taken a very different path. Community-trained models often show improved accuracy on popular benchmarks such as MATH or AIME24~\citep{liu2025understanding,song2025outcome,chen2025pass,cheng2025reasoning,cui2025entropy,shao2025spurious,wang2025reinforcement,yu2025dapo}. However, these RL models typically succeed only on problems that their corresponding base models could already solve given realistic sampling budgets~\citep{wu2025invisibleleashrlvrescape,yue2025doesreinforcementlearningreally}. This is a far cry from earlier RL successes, where base policies were incapable of solving tasks outright and progress required genuine exploration and skill acquisition. This disconnect between RL's exploratory promise and its current application reflects a substantial blindspot in the current open-source evaluation ecosystem. Because several open-source base models already achieve nearly 100\% \texttt{pass@1024} on several popular benchmarks~\citep{yue2025doesreinforcementlearningreally}, test sets in their current form are fundamentally inadequate for measuring---or encouraging---genuine progress in reasoning beyond the base model's reach.

To address this gap, we introduce \textbf{MATH-Beyond (\textsc{MATH-B})}, a new benchmark of high-school–level competition math problems, specifically constructed so that popular open-weight base models are unlikely to solve even with 1024 attempts. As a result, progress on \textsc{MATH-B} necessarily requires expanding the reasoning capabilities of base models, making it an ideal target for academic research.

\textsc{MATH-B} is constructed by filtering mathematical reasoning datasets (DAPO-Math-17K~\citep{yu2025dapo} and DeepScaleR~\citep{deepscaler2025}), resulting in problems that are topically indistinguishable from those in standard benchmarks. While constructing the dataset, we also uncovered and addressed several non-obvious failure modes in programmatic verification, which informed our final benchmark design. To ensure correctness, all problems are additionally verified against stronger reasoning models such as GPT-5-Mini or o4-mini-high, which reliably solve them. We further confirm that leading community RL models, including Nemotron-Research-Reasoning-Qwen-1.5B~\citep{liu2025prorlprolongedreinforcementlearning} and DeepScaleR-1.5B, perform poorly on \textsc{MATH-B}, underscoring the limitations of current approaches and the need for methods that extend reasoning capabilities. In summary, our contributions are as follows:

\begin{itemize}
    \item \textbf{New benchmark suite:} We construct MATH-Beyond, a benchmark suite derived from the failures of a large and diverse set of base models (\texttt{pass@1024} $\approx 0$). This suite includes: a union set of 181 problems unsolved by at least one of these models; model-specific benchmarks for targeted analysis; and a highly challenging intersection set of 41 problems that proved unsolvable for the entire considered set. To ensure quality, all problems are annotated for topic and difficulty following the procedure from Omni-MATH~\citep{gao2024omnimathuniversalolympiadlevel}, and undergo answer verification by frontier models (o4-mini-high \& GPT5-Mini).

    \item \textbf{Verification pitfalls:} During benchmark construction, we observed several pitfalls in standard RLVR verification. We take these into account in our benchmark design and also document them, highlighting subtle edge cases for the community to be aware of.
    
    \item \textbf{Model evaluation:} We evaluate RL-finetuned models such as Nemotron-Research-Reasoning-Qwen-1.5B, DeepScaleR-1.5B, and Skywork-OR1~\citep{he2025skyworkopenreasoner1} on MATH-B and find that they do not substantially expand reasoning boundaries. In contrast, we find that newer model families like  Qwen3-4B and Qwen3-8B~\citep{yang2025qwen3technicalreport} perform better, presumably due to better distributional overlap with our dataset.
\end{itemize}

\section{Framework for Evaluating Model Expansion}\label{sec:theory}

Our goal is to quantify whether a post-trained model (e.g., after RLVR) \emph{expands} its reasoning capabilities beyond its base model. While post-training can affect many aspects (e.g., robustness, exploration, or transfer), our framework is designed to isolate the specific phenomenon of \emph{boundary expansion}—what new problems the post-trained model can solve relative to its base model. This section formalizes this evaluation framework, adapting the nomenclature and definitions from \cite{wu2025invisibleleashrlvrescape} into a slightly simplified, empirical version. We apply this framework to our benchmark \textsc{MATH-B}, which is a \textit{``zero-baseline''} test. We define this as a setting where a benchmark is specifically constructed such that the base model empirically fails on all problems (i.e., its observed \texttt{pass@k} is zero) given a realistic sampling budget.

\subsection{The Empirical \texttt{pass@k} Metric}

We evaluate a post-trained policy $\pi$ against its base model $q$ on a dataset $D$. Our evaluation is based on the empirical \texttt{pass@k} metric. For a given problem $x \in D$ and a policy $p \in \{\pi, q\}$ (usually an LLM), we draw $k$ i.i.d.\ samples $\{y_1, \dots, y_k\}$. Let $\mathcal{C}(x)$ be the set of all correct completions for $x$. The empirical success is:
\[
\texttt{pass@k}(p;x) =
\begin{cases}
1 & \text{if } \exists i \in \{1,\dots,k\} \text{ such that } y_i \in \mathcal{C}(x), \\[4pt]
0 & \text{otherwise}.
\end{cases}
\]
This binary value---1 for a ``pass'' and 0 for a ``fail''---is the ground-truth measure of success for a single problem.

With a slight abuse of notation, the central metric reported in our paper is the average success rate across the entire dataset $D$:
\[
\texttt{pass@k}(p) = \frac{1}{|D|} \sum_{x \in D} \texttt{pass@k}(p;x),
\]
the \textbf{ratio of problems in $D$ solved by policy $p$} using a $k$-sample budget.

\subsection{Decomposition and Key Metrics}\label{subsec:metrics}

To understand \emph{how} $\pi$'s performance differs from $q$'s, we first define the \textbf{Reachable Set} $\mathcal{R}_k(p, D)$ as the set of problems $p$ solves:
\[
\mathcal{R}_k(p, D) = \{ x \in D : \texttt{pass@k}(p;x) = 1 \}.
\]
Comparing the reachable sets of $\pi$ and $q$ allows us to isolate our primary metric and contextualize it.

\paragraph{Expansion (Primary Metric).}
Our \emph{primary focus} and sole reported metric is the \textbf{Expansion Rate}. This measures genuine boundary expansion—new problems $\pi$ solves that $q$ could not. It is defined based on the \textbf{Expansion Set} $\mathcal{E}_k = \mathcal{R}_k(\pi, D) \setminus \mathcal{R}_k(q, D)$:
\[
\text{\textbf{Expansion Rate}} = \frac{|\mathcal{E}_k|}{|D|}.
\]

\paragraph{Shrinkage.}
For diagnostic context, we also define \textit{Shrinkage} (or ``forgetting''). This measures problems $q$ could solve that $\pi$ now fails, defined by the \textit{Shrinkage Set} $\mathcal{S}_k = \mathcal{R}_k(q, D) \setminus \mathcal{R}_k(\pi, D)$. It can be quantified as:
\[
\text{Shrinkage Rate} = \frac{|\mathcal{S}_k|}{|D|}.
\]

\paragraph{Preservation.}
Similarly, \textit{Preservation} measures the fraction of $q$'s capabilities that $\pi$ retains, defined by the \textit{Preservation Set} $\mathcal{P}_k = \mathcal{R}_k(\pi, D) \cap \mathcal{R}_k(q, D)$. It is quantified as:
\[
\text{Preservation Rate} = \frac{|\mathcal{P}_k|}{|\mathcal{R}_k(q, D)|}.
\]

\paragraph{Consolidation.}
Finally, \textit{Consolidation} is a concept for measuring if preserved solutions become more robust (i.e., solvable at $\texttt{pass@1}$). It is defined as:
\[
C_k(\pi, q) = \frac{|\mathcal{P}_k \cap \mathcal{R}_1(\pi, D)|}{|\mathcal{P}_k|}.
\]

\paragraph{Interpretation.}
The overall pass rate of $\pi$ decomposes as $\texttt{pass@k}(\pi) = (|\mathcal{E}_k| + |\mathcal{P}_k|) / |D|$. While Shrinkage, Preservation, and Consolidation are crucial concepts for a full diagnosis, our target for \emph{expanding the reasoning boundary} is squarely the \textbf{Expansion Rate}. The other concepts serve as a theoretical guardrail to ensure that measured gains are from genuine expansion, not mere reallocation.

\subsection{Special Case: The \textsc{MATH-B} Benchmark}\label{subsec:mathb}

The metrics framework in Section~\ref{subsec:metrics} simplifies cleanly for \textsc{MATH-B}. This benchmark is constructed as a ``zero-baseline'' meaning it is composed of problems where the base model $q$ was empirically observed to fail within the sampling budget $k$ (i.e. \texttt{pass@k}  $\approx 0$). In our evaluation, this means the base model's reachable set is empty::
\[
\mathcal{R}_k(q, D) = \varnothing.
\]
Under this premise, the decomposition from Section~\ref{subsec:metrics} collapses:
\[
\mathcal{S}_k = \varnothing, \qquad \mathcal{P}_k = \varnothing, \qquad \mathcal{E}_k = \mathcal{R}_k(\pi, D).
\]
Shrinkage and preservation are thus not applicable. Every solve by $\pi$ is by definition an expansion, and the single relevant metric becomes:
\[
\text{Expansion Rate} = \frac{|\mathcal{R}_k(\pi, D)|}{|D|} = \texttt{pass@k}(\pi).
\]
A ``win'' on \textsc{MATH-B} is simply a positive Expansion Rate, providing an unambiguous readout of genuine boundary expansion.

\section{Creating the benchmark}\label{sec:dataset}
\subsection{On the Pitfalls of Verification}\label{sec:verification_issues}

The goal of our work is to identify questions that models fail at high \texttt{pass@1024} due to genuine methodological or calculation errors, not because of artifacts in the verification pipeline. Current rule-based verification methods for math, used both in training and evaluation, are themselves prone to systematic failures. These failures can mask true performance by penalizing answers where the reasoning or ground truth is technically correct but formatted poorly, making it appear as though the model failed when the verifier simply could not handle the formatting or parsing.

To guide our evaluation and benchmark design, and to alert the community to these pitfalls, we sketch seven distinct failure modes observed across common RL finetuning and evaluation frameworks. We uncovered these issues through an analysis of reasoning traces produced by DeepSeek-R1-Distill-Qwen2.5-7B~\citep{deepseekai2025deepseekr1incentivizingreasoningcapability} when applied to a subset of NuminaMath-1.5~\citep{numina_math_datasets}. Table~\ref{tab:failure-modes} provides concise but complete examples (problem, ground truth, model output snippet, and the precise mismatch), and Table~\ref{tab:all-frameworks-comparison} in the Appendix summarizes how different frameworks are affected. 

\begin{table}[ht]
\centering
\small
\renewcommand{\arraystretch}{1.15}
\captionsetup{skip=5pt} 
\caption{\textbf{Common failure modes in rule-based math answer verification}. These failures often stem from rigid heuristics, such as reading only the first or last boxed answer, requiring specific text anchors (e.g., "Answer:"), or other parsing failures. Each row shows the ground truth (GT), a model snippet, and the resulting verifier error.}
\label{tab:failure-modes}
\begin{tabularx}{\columnwidth}{l p{0.1\columnwidth} p{0.30\columnwidth} X}
\toprule
\textbf{Mode} & \textbf{GT} & \textbf{Model snippet} & \textbf{Verifier behavior} \\
\midrule
\textbf{F1: Multiple valid} &
\texttt{5 or 13} &
… $x=\boxed{5}$ or $\boxed{13}$ &
\emph{Expected:} accept both values. \emph{Actual:} only first/last boxed read; one valid ignored. \\
\textbf{F2: Late correct} &
\texttt{150} &
… $\boxed{50}$; … $\boxed{100}$; … $\boxed{150}$ &
\emph{Expected:} final value only. \emph{Actual:} early intermediate captured. \\
\textbf{F3: Early correct} &
\texttt{42} &
… $\boxed{42}$ … later $\boxed{84}$ &
\emph{Expected:} accept any correct boxed answer. \emph{Actual:} later wrong overrides early correct. \\
\textbf{F4: Corrections} &
\texttt{25} &
… $\boxed{20}$, then correcting $\boxed{25}$ &
\emph{Expected:} use corrected value. \emph{Actual:} first match persists; “last-only” works, “first-only” fails. \\
\textbf{F5: Unordered tuple} &
\texttt{3,4,5} &
… $\boxed{5}$, $\boxed{3}$, $\boxed{4}$ &
\emph{Expected:} order-agnostic match. \emph{Actual:} order-sensitive; heuristics misgrade. \\
\textbf{F6: Missing anchors} &
\texttt{4} &
… $2+2=\boxed{4}$ (no ``Answer:'' prefix) &
\emph{Expected:} accept boxed numeral alone. \emph{Actual:} anchor required; otherwise wrong. \\
\textbf{F7: MCQ partial} &
\texttt{C)1000} &
… ``The answer is $\boxed{C}$.'' &
\emph{Expected:} accept label-only (\texttt{C}). \emph{Actual:} requires exact ``C)1000''. \\
\bottomrule
\end{tabularx}
\end{table}

\subsection{MATH-B Construction Pipeline}\label{sec:dataset_creation}

Our benchmark is the result of a multi-stage filtering pipeline designed to isolate problems that are (a) correct and unambiguous, (b) novel, and (c) demonstrably unsolvable by a suite of open-source base models even at a high sampling budget. The process is detailed below and in \Cref{fig:schematic}. 

\subsubsection{Base Dataset Sourcing}
Our construction begins by sourcing a large pool of candidate problems. The primary goal, as stated, is to find problems that defeat open-source models at high sampling budgets (\texttt{pass@1024}). This requirement immediately disqualifies many common datasets. For instance, DeepMath103K~\citep{he2025deepmath103klargescalechallengingdecontaminated} and Big-Math~\citep{albalak2025bigmathlargescalehighqualitymath} are explicitly designed to be solvable (e.g., for GRPO training), and indeed, models like Llama-3.1-8B solve them with few attempts. We also sought to minimize data contamination, ruling out datasets like NuminaMath which have been seen in pretraining by models like Qwen2.5-Math~\citep{yang2024qwen25mathtechnicalreportmathematical} and consequently DeepSeek-R1.

We therefore selected two base datasets: \textbf{DAPO-Math-17K}~\citep{yu2025dapoopensourcellmreinforcement}, which satisfies our criteria for both high difficulty (not verified by open-source models) and novelty (released after DeepsSeek-R1), and \textbf{DeepScaleR}~\citep{deepscaler2025}, which, while potentially seen, provides a large corpus of problems that also lack open-source verification. This combined set forms our initial candidate pool of 53,682 problems.

\subsubsection{Quality filters}\label{sec:quality_filters}

We then apply an array of quality filters as described in the following paragraphs.

\paragraph{Question and Answer-Type Filtering}
To mitigate the verification pitfalls detailed in Section~\ref{sec:verification_issues}, we apply several deterministic filters. First, to avoid ambiguity from failure modes like F1 (Multiple valid answers), and F5 (Unordered Tuples), we filter DeepScaleR to include only problems with integer-based ground truths (DAPO-Math-17K already meets this criterion). Second, using regex-based filters, we remove all multiple-choice questions (to prevent F7 parsing failures) and any questions containing Chinese characters. Finally, to ensure all problems are self-contained, we remove any questions referencing external figures or images.
This filtering process reduced the pool to 34,515 datapoints.

\paragraph{Pre-screening and Random Sampling}
The problems from the previous step still represent a computationally prohibitively large set for large-scale evaluation. To reduce this pool, we first perform a difficulty pre-screening step. We evaluate Deepseek-R1-Distill-Qwen2.5-7B on all problems, keeping only those that it \emph{could not solve} within a \texttt{pass@16} budget. In addition to the screening, we randomly sample a portion of this dataset for further processing.

\paragraph{Ground-Truth Answer Verification}
A critical step is to ensure that these problems are unsolved due to their intrinsic challenge, not because their provided ground-truth answers are incorrect. While our source datasets are generally reliable (e.g., DeepScaleR is derived from AIME~\citep{MAA_AIME} and AMC~\citep{MAA_AMC_2023} competitions~\citep{deepscaler2025}), we took an additional step to ensure the ground-truth answers are correct. We evaluate (\texttt{pass@2}) this pre-screened subset using two frontier-class models, o4-mini-high and GPT-5-Mini\citep{balunovic_srimatharena_2025}. For each problem, we prompt the models to \textit{"Think step-by-step and put your final answer in \texttt{\textbackslash boxed{}}."} and check if the extracted answer matches the dataset's ground truth. We retain only the problems where \emph{at least one} of these frontier models successfully reproduced the ground-truth answer.

\paragraph{Deduplication Against Standard Benchmarks}
To ensure the novelty of our dataset, we first perform an exact string-match deduplication of our candidate problems against several common test sets, including MATH-500~\citep{hendrycks2021measuringmathematicalproblemsolving}, MinervaMath~\citep{lewkowycz2022solvingquantitativereasoningproblems}, OlympiadBench~\citep{he2024olympiadbenchchallengingbenchmarkpromoting}, AMC23~\citep{MAA_AMC_2023}, AIME-2024, and AIME-2025~\citep{MAA_AIME}. We confirmed that no question from our set is present in these benchmarks. At the end of this stage, a collection of 184 problems remain.

\subsubsection{Final Benchmark Construction (\texttt{pass@1024} Filtering)}\label{sec:final_pass1k_filtering}

The final step in our benchmark's construction is to filter the candidate problems using a diverse suite of representative open-source models. The definition of a `\textit{base model}' is contextual; it typically refers to a model intended for further fine-tuning. Keeping in mind the models the community often uses for post-training research, we select a suite of models categorized as either `base' or `supplementary' models. As we will detail, these two groups are used to construct different subsets of our final benchmark.

\paragraph{Base Models} This set is used to define the most challenging subset of our benchmark. It includes: Qwen2.5-1.5B, Qwen2.5-7B~\citep{qwen2025qwen25technicalreport}, Qwen2.5-Math-1.5B, Qwen2.5-Math-7B~\citep{yang2024qwen25mathtechnicalreportmathematical}, Qwen3-4B-Base, Qwen3-8B-Base~\citep{yang2025qwen3technicalreport}, DeepSeek-R1-Distill-Qwen2.5-1.5B, DeepSeek-R1-Distill-Qwen2.5-7B~\citep{deepseekai2025deepseekr1incentivizingreasoningcapability}, OLMo-7B~\citep{groeneveld2024olmoacceleratingsciencelanguage}, OLMo-2-7B~\citep{olmo20252olmo2furious}, and Llama-3.1-8B~\citep{grattafiori2024llama3herdmodels}.

\paragraph{Supplementary Models} This group is combined with the base models to define the full benchmark. It includes: Qwen2.5-1.5B-Instruct, Qwen2.5-7B-Instruct~\citep{qwen2025qwen25technicalreport}, Qwen2.5-Math-1.5B-Instruct, Qwen2.5-Math-7B-Instruct~\citep{yang2024qwen25mathtechnicalreportmathematical}, Qwen3-4B, Qwen3-8B~\citep{yang2025qwen3technicalreport}, DeepScaler-1.5B~\citep{deepscaler2025}, Nemotron-Research-Reasoning-Qwen-1.5B(v1 and v2)~\citep{liu2025prorlprolongedreinforcementlearning}, and Skywork-OR1-7B~\citep{he2025skyworkopenreasoner1}.

The deduplicated and pre-screened candidate set is then subjected to our final filtering stage. We evaluate the problems against \emph{all listed models} generating 1024 samples for each. During this evaluation, we apply our robust verification logic (see~\Cref{tab:failure-modes} for failure modes) to ensure we were measuring genuine reasoning failures. Our experiments and analyses required over 20\,000 A100 GPU hours. Please refer to \Cref{sec:appx_inference} for more details on the evaluation parameters.

This comprehensive evaluation allows us to define our final benchmark, \textbf{MATH-B}, which comprises three broad subsets for targeted analysis:
\begin{itemize}
    \item \textbf{MATH-Beyond-Union Set (MATH-B-U):} The full benchmark of \textbf{181 problems}, containing any problem that at least one model from our \emph{entire suite} (both base and post-trained) failed to solve within 1024 attempts.
    \item \textbf{MATH-Beyond-Intersection Set (MATH-B-I):} A more challenging core subset of \textbf{41 problems}, that \emph{all} of our considered \textbf{base models} failed to solve. This is a hard subset of MATH-B-U (see \Cref{tab:questions-intersection} for the QA pairs).
    \item \textbf{Model-Specific Sets:} For any given model in our suite, this is the collection of all problems in the Union Set that it failed to solve (see \Cref{tab:unsolved-pass1024-all}). These sets enable fine-grained testing of RL finetuned models derived from one of the evaluated base models.
\end{itemize}
Overall, our MATH-B datasets serve as benchmarks of reasoning capability expansion (see~\Cref{sec:theory}).

\subsection{Data Characteristics and Key Remarks}\label{sec:base_models}

\begin{figure}[h]
    \centering
    \includegraphics[width=1.0\linewidth]{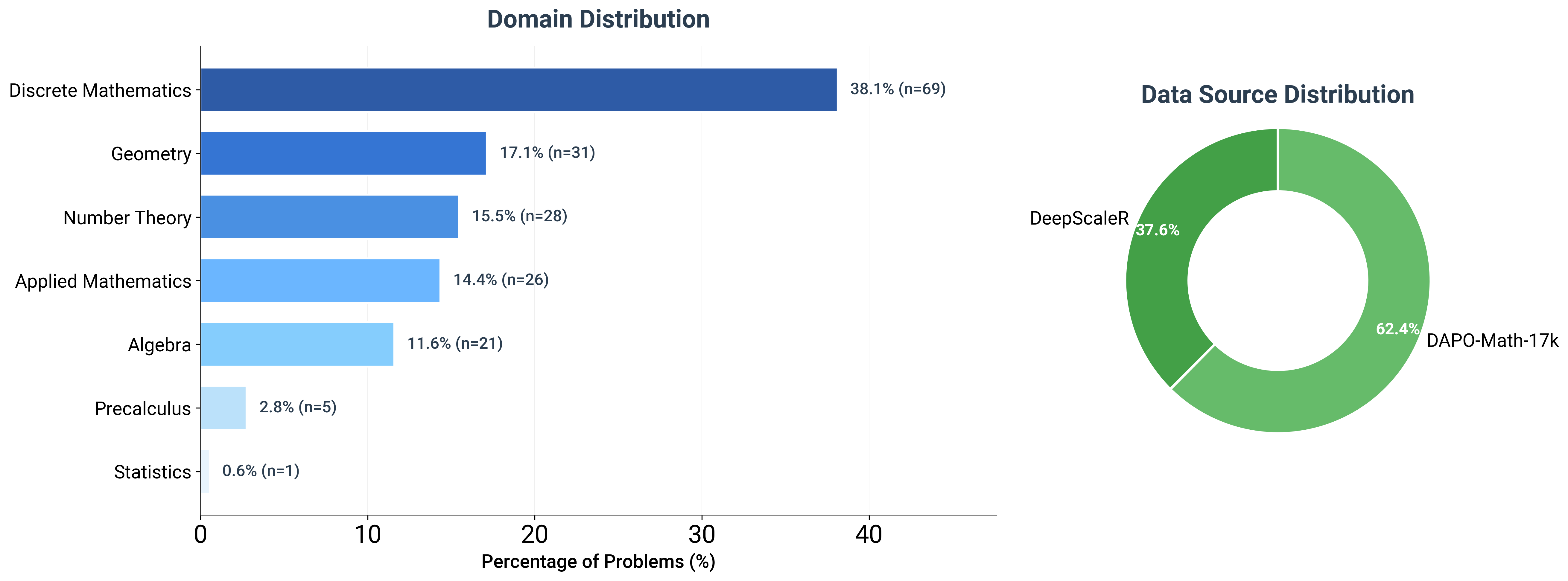}
    \caption{\textbf{Characteristics of the MATH-B-U dataset}. Left subplot shows the distribution of math domains. Right subplot show the distribution of source datasets.}
    \label{fig:domain_source}
\end{figure}

\paragraph{Domain and Difficulty Annotation}
To analyze the characteristics of the MATH-B dataset, we annotate each problem for its mathematical domain and human-perceived difficulty. We adapt the procedure from Omni-MATH~\citep{gao2024omnimathuniversalolympiadlevel}, using GPT-5 to perform the labeling based on a contrastive prompting strategy (see supplementary material for the full prompt). This method leverages labeled examples from various math datasets to contextualize and assign scores to new problems and has shown to be aligned well with human judgment~\citep{gao2024omnimathuniversalolympiadlevel}. A selection of annotated problems from MATH-B-I is provided in \Cref{tab:questions-intersection}. We also plot the distribution of ground truth answers in \Cref{fig:final_answer}.

\begin{figure}[h]
    \centering
    \includegraphics[width=1.0\linewidth]{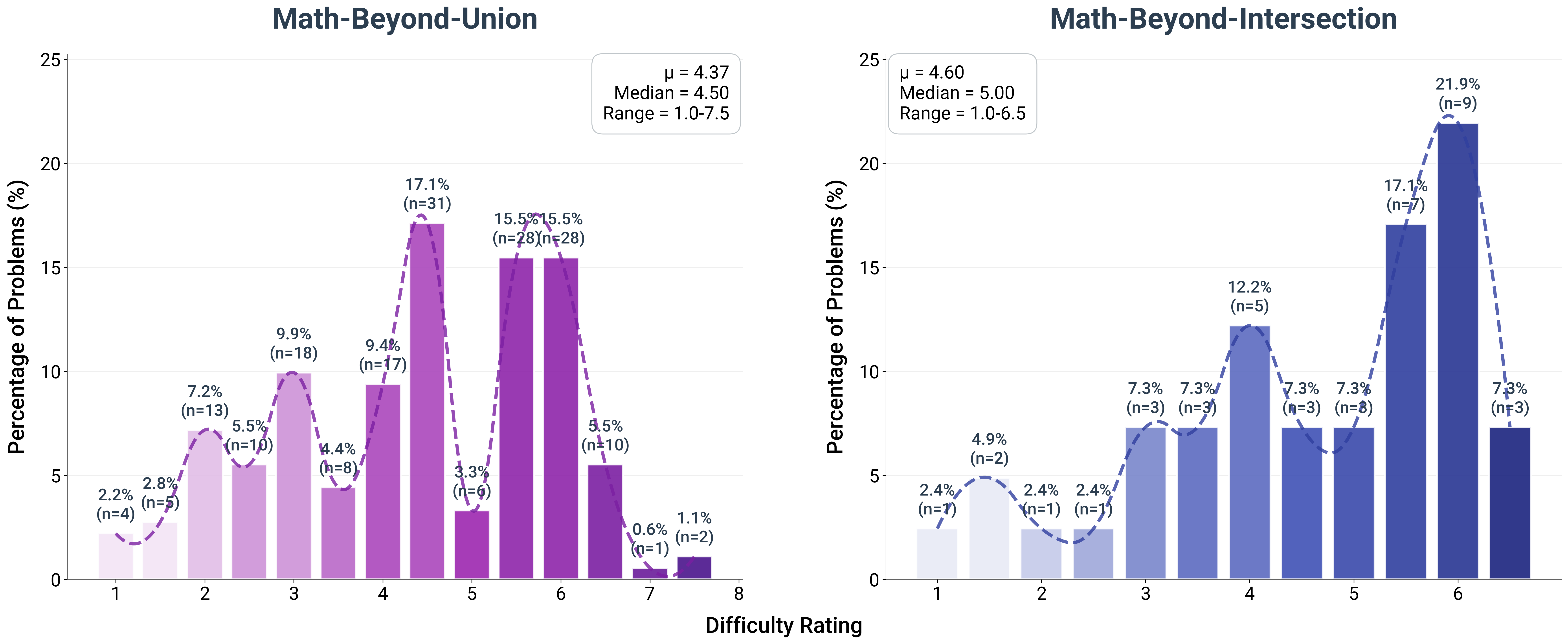}
    \caption{\textbf{Difficulty distribution.} The left subplot shows the difficulty distribution of the MATH-Beyond-Union set, while the right subplot shows that of the MATH-Beyond-Intersection set. The wide spread of difficulty levels highlights a key mismatch: the problems that models find challenging are not necessarily those that humans typically struggle with.}
    \label{fig:difficulty}
\end{figure} 

\paragraph{Distribution Analysis}
As shown in~\Cref{fig:domain_source}, the topic distribution of MATH-B consists entirely of standard high-school mathematics subjects, ensuring topical relevance. The difficulty distribution, plotted in~\Cref{fig:difficulty}, reveals a wide range for both the Union and Intersection sets, with a median human-difficulty rating of 4 out of 10. Notably, even for the challenging MATH-B Intersection set, the maximum difficulty score is only 6.5. This suggests a significant disconnect between human-perceived difficulty and model failure modes; problems that are not considered exceptionally hard for humans can still be robustly unsolvable by current models.

\paragraph{Benchmark Justification}
A strong benchmark should be realistic, difficult, and efficient~\citep{nie2025uqassessinglanguagemodels}. \textsc{MATH-B} meets these with (i) \emph{realism}: problems drawn from standard high-school curricula; (ii) \emph{difficulty}: base models perform poorly even with a large sampling budget (i.e. \texttt{pass@1024}); and (iii) \emph{efficiency}: the small suite enables potentially rapid, low-cost evaluation, and—as shown in \Cref{sec:evaluation}—post-trained models exhibit only marginal gains at \texttt{pass@1024}, making \textsc{MATH-B} a perfect test bed for research.

\paragraph{Benchmark Usage}
We expect researchers to use \textsc{MATH-B} with the base models and their corresponding splits specified in \Cref{sec:final_pass1k_filtering}: (1) evaluate the base model to estimate its \texttt{pass@k} for reasonably large $k$ (e.g., $k=1024$, expected to be nearly 0); (2) apply some RL method of interest; (3) re-estimate \texttt{pass@k} for the post-trained policy, which indicates the \emph{Expansion Rate} per \Cref{sec:theory}. The benchmark is intended specifically for methods that aim to \emph{expand} the listed base model’s reasoning boundary. While this is a research benchmark, we hope the methods that people come up with scales to larger scales and other scenarios

\section{Evaluating Expansion Across Finetuning Methods}
\label{sec:evaluation}

\begin{figure}[h!]
    \centering
    \includegraphics[width=0.8\linewidth]{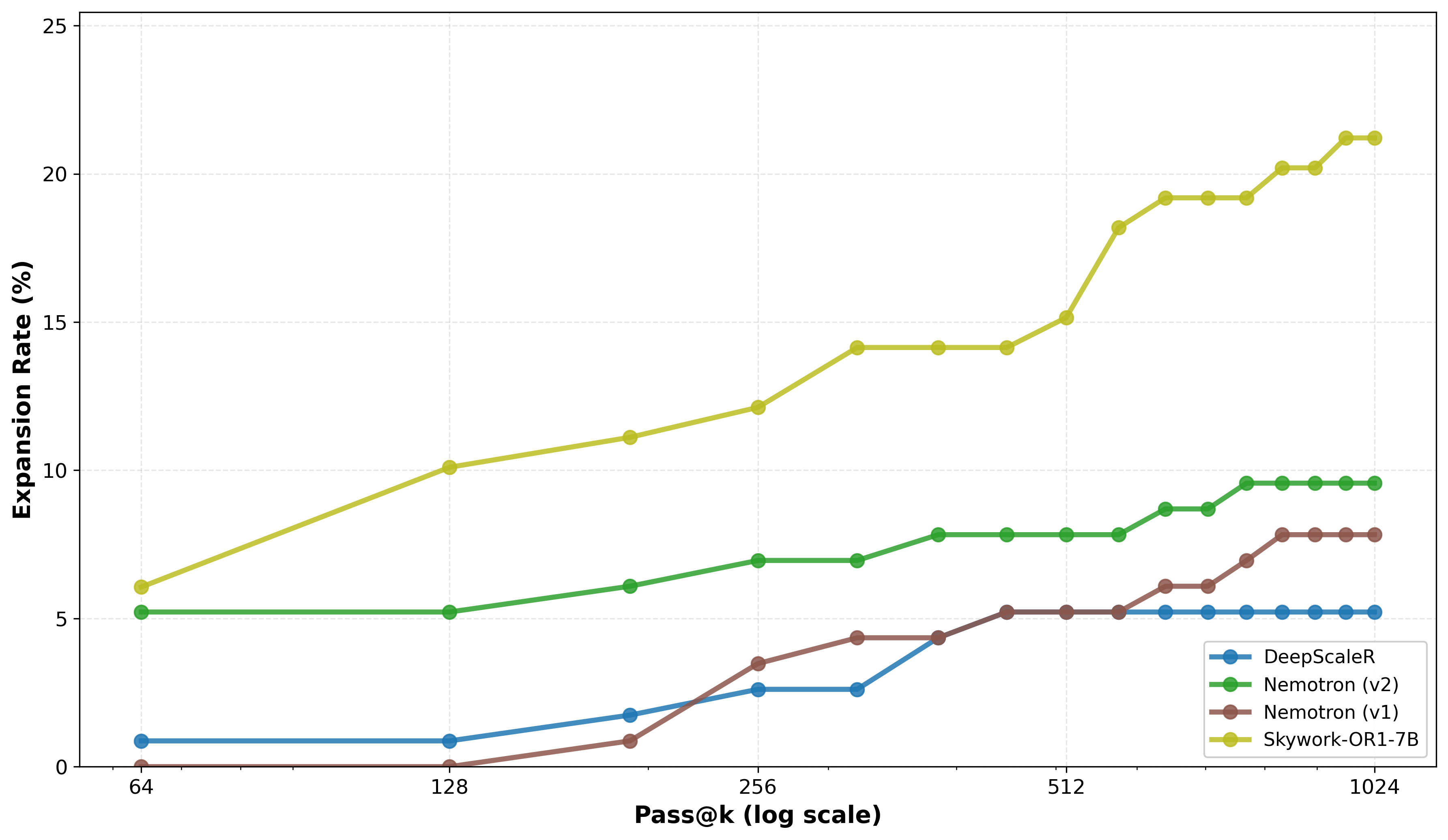}
    \caption{\textbf{Evolution of Expansion Rate for RL Models.} Models are evaluated on the \textsc{MATH-B} problems failed by their respective base models (115 for R1-Qwen2.5-1.5B; 99 for R1-Qwen2.5-7B).}
    \label{fig:exp_rate_evol}
\end{figure}

We evaluate several post-trained models on MATH-B to measure their ability to expand beyond their base model's reasoning capabilities (i.e. Expansion Rate~\Cref{sec:theory}). Our analysis of models finetuned with reinforcement learning (RL) reveals that current methods achieve only modest expansion. As shown in Table~\ref{tab:expansion_rates}, the three RL models based on DeepSeek-R1-Distill-Qwen1.5B (r1-1.5b)~\citep{deepseekai2025deepseekr1incentivizingreasoningcapability} solve fewer than 10\% of the test problems. In contrast, Skywork-OR1-7B (skywork\_or1)~\citep{he2025skyworkopenreasoner1} reaches a more promising 21\% expansion; notably, its training involves adaptive entropy control and a higher temperature, likely affording a greater scope for exploration. Keep in mind that these are \texttt{pass@1024} evaluations. This result suggests that RL techniques designed to explicitly encourage exploration may indeed result in higher Expansion Rates. We also observe that while prolonged training (i.e. more RL compute) can offer marginal gains---as seen in the slight improvement from Nemotron-Research-Reasoning-Qwen-1.5B version 1 (nemotron\_v1) to Nemotron-Research-Reasoning-Qwen-1.5B version 2 (nemotron\_v2)~\citep{liu2025prorlprolongedreinforcementlearning}---the rather small ~1.5\% increase underscores the need for more efficient and effective exploration methods. See also the evolution of Expansion Rate in \Cref{fig:exp_rate_evol}.

\begin{table}[h!]
\centering
\small
\setlength{\tabcolsep}{6pt}
\caption{\textbf{Expansion Rates of post-trained models} using either Reinforcement Learning (RL) or Supervised Fine-Tuning (SFT) /  Distillation. The Expansion Rate measures the percentage of previously unsolvable problems (from the base model's perspective) that the post-trained model can now solve at \texttt{pass@1024}.}
\label{tab:expansion_rates}
\begin{tabularx}{\columnwidth}{@{}l l l r r@{}}
\toprule
\textbf{Base model} & \textbf{Post-trained model} & \textbf{Method} & \textbf{Base unsolved} & \textbf{Expansion Rate (\%)} \\
\midrule
\multicolumn{5}{@{}l}{\textit{Reinforcement Learning (RL) models}} \\
r1-1.5b& nemotron\_v1 & RL & 115 & 7.83 \\
r1-1.5b & nemotron\_v2 & RL & 115 & 9.57 \\
r1-1.5b & DeepScaleR & RL & 115 & 5.22 \\
\addlinespace[2pt]
r1-7b & skywork\_or1 & RL & 99 & 21.21 \\
\midrule
\multicolumn{5}{@{}l}{\textit{Supervised Fine-Tuning (SFT) models}} \\
Qwen3-4B-base & Qwen3-4B & Long CoT Dist. & 112 & 58.93 \\
Qwen3-8B-base & Qwen3-8B & Long CoT Dist. & 116 & 66.38 \\
\bottomrule
\end{tabularx}
\end{table}

As an illustrative contrast, we evaluated Qwen3-4B and Qwen3-8B. These models are the result of finetuning their respective base versions (Qwen3-4B-Base and Qwen3-8B-Base) by distilling long Chain-of-Thought (CoT) reasoning trajectories from a more capable teacher model. They demonstrate substantially higher Expansion Rates of 58.93\% and 66.38\%, respectively. While this is not a direct apples-to-apples comparison due to differing training setups, the result is highly informative: it shows that significant expansion is achievable when a model is exposed to the correct distribution of reasoning steps, an overlap that long CoT distillation provides~\citep{yang2025qwen3technicalreport}.

This contrast highlights that the primary limitation of current RL techniques may not be an inherent inability of the base models to learn, but rather the failure of the exploration process to find these effective reasoning pathways on its own. Developing RL methods that can discover these pathways without a teacher model remains a key challenge for reaching frontier capabilities and is the central motivation for our work.

\section{Discussion and Related Work}
\label{sec:disc_rel}

\paragraph{On the Choice of \textbf{\textit{k}} $ = \mathbf{1024}$}
Our selection of $k=1024$ for all \texttt{pass@k} evaluations is a deliberate choice to ensure our benchmark is challenging, stable, and efficient. Firstly, a large sampling budget is necessary to probe the true reasoning boundary of a model. Many popular benchmarks become saturated at this scale, with base models solving a high percentage of problems and leaving no room to measure improvement~\citep{yue2025doesreinforcementlearningreally}. We chose $k=1024$ precisely because it represents a budget where \textsc{MATH-B} remains difficult, creating a meaningful testbed for genuine expansion. Secondly, this choice is strongly supported by our empirical analysis. While overall \texttt{pass@k} performance shows a consistent log-linear increase with the sampling budget (\Cref{fig:all_models_passk_curves}), the marginal gains for each additional sample diminish considerably (\Cref{fig:gains_per_k}). Most importantly, the Expansion Rate for RL-finetuned models—our core metric for progress—largely plateaus as the budget approaches 1024 (\Cref{fig:exp_rate_evol}). Therefore, $k=1024$ represents a principled trade-off: it is large enough to push models beyond their comfort zone on a challenging benchmark, yet provides a stable and computationally feasible point to reliably measure the expansion of the reasoning boundary.

\paragraph{Limitations of Existing Benchmarks and Metrics}
In mathematical reasoning, progress is often measured by \texttt{pass@k}. For $k>1$, this metric is taken to be indicative of a model's exploratory potential. However, \texttt{pass@k} is an incomplete measure of exploratory potential, as it conflates the sharpening of existing solutions (Consolidation) with the discovery of entirely new ones (Expansion)~\citep{wu2025invisibleleashrlvrescape}. As detailed in \Cref{sec:theory}, our work is concerned with Expansion. Many existing benchmarks are now saturated by strong open-source base models, making it impossible to measure new boundary expansion~\citep{yue2025doesreinforcementlearningreally,balunovic_srimatharena_2025,wu2025invisibleleashrlvrescape}. Furthermore, these benchmarks are targets of hyper-optimization, potentially rewarding spurious correlations~\citep{shao2025spurious}. \textsc{MATH-B} addresses this by serving as a diagnostic tool. It comprises problems that are topically standard high-school math but are constructed to expose the subtle failures and brittleness of the dominant open-source research paradigm.

\paragraph{Our Empirical Contribution in Context}
Building on the framework of \citet{wu2025invisibleleashrlvrescape}, we \emph{instantiate} these ideas in a concrete, reusable benchmark. their work introduces and analyzes Expansion within a specific setting, whereas we put forth a zero-baseline dataset (i.e., \texttt{pass@1024}\,$\approx 0$ for the base models) that operationalizes the concept and scales evaluation across a wide range of open-weight models. This offers a practical path to shift community focus from merely improving \texttt{pass@k} on saturated suites to \emph{demonstrably expanding the reasoning boundary}. Further, while \textsc{MATH-B} is instantiated for a specific set of open-weight base models, we expect a subset of items to remain zero-baseline for larger open-weight models. We did not evaluate those due to resource constraints. Since MATH-B is intended to drive RL methods that \emph{expand} a given base model’s reasoning boundary (rather than chase model-specific quirks), we expect the resulting methods and insights to transfer across model families and scales.

\paragraph{Parallels to OOD and Adversarial Benchmarking}
Our benchmark's design parallels adversarial and out-of-distribution (OOD) evaluations in computer vision, such as ImageNet-Adversarial~\citep{hendrycks2021naturaladversarialexamples} or ImageNet-Sketch~\citep{wang2019learningrobustglobalrepresentations}. These benchmarks revealed that models were brittle to inputs that vary mildly from the training data~\citep{mayilvahanan2024doesclipsgeneralizationperformance,mayilvahanan2025searchforgottendomaingeneralization}. They showed that these models have a distributional generalization quality to them~\citep{fang2022datadeterminesdistributionalrobustness,mayilvahanan2025llmslinedatadetermines}. \textsc{MATH-B} functions as an OOD test for reasoning: the problems are topically in-distribution, but their solutions are effectively out-of-distribution for base models. It thereby stress-tests the brittleness of the current (base model + GRPO-style) setup, highlighting the absence of genuine exploration.

\paragraph{A Testbed for Novel Reasoning Paths}
This benchmark provides an empirical tool to quantify the "discovery vs. sharpening" debate. Success on \textsc{MATH-B} requires more than re-weighting existing solution modes; it necessitates finding solution paths that are effectively unreachable by the base model. For a post-trained model to solve these problems, its generator-verifier system must produce and reward solution distributions that are qualitatively different from the base model's initial distribution. The challenge we pose, therefore, is for new methods to find these novel reasoning paths. Many in the community identify focusing on exploration as the way forward~\citep{liu2025understanding,song2025outcome,chen2025pass,cheng2025reasoning,cui2025entropy,shao2025spurious,wang2025reinforcement,yu2025dapo}. However, progress is often still measured with \texttt{pass@k} on saturated benchmarks with small $k$ values and without isolating for Expansion. We contend that progress will accelerate by developing novel exploration methods and evaluating Expansion Rate on challenging benchmarks like ours.

\section{Conclusion}
\label{sec:conclusion}
Our findings indicate that current post-training methods in the open-source ecosystem, particularly for models up to 8B parameters, primarily refine pre-existing reasoning abilities rather than creating new ones. The poor performance of these post-trained models on \textsc{MATH-B} empirically confirms that they struggle to expand their capabilities to problems that lie just beyond their base model's reach. We introduce \textsc{MATH-B} as a precise diagnostic tool to address this issue. Its purpose is to catalyze research into post-training methods that achieve genuine exploration, providing a clear and reliable signal for when a model has truly expanded the boundaries of machine reasoning.

\newpage
\section{Reproducibility Statement}
We are committed to ensuring the reproducibility of our work. All code required to reproduce the
experiments is provided in the supplementary material. Detailed derivations are
included in the main text (\Cref{sec:theory}). Experimental settings and hyperparameters are described in the main paper (\Cref{sec:dataset_creation})
and supplementary sections (\Cref{sec:appx_inference}). Our data is available at \url{https://huggingface.co/datasets/brendel-group/MATH-Beyond}. Our code is available at \url{https://brendel-group.github.io/math-beyond/}.
\bibliography{iclr2026_conference}
\bibliographystyle{iclr2026_conference}

\newpage
\appendix
\section{Appendix}
\subsection{Verification issues in different frameworks}\label{sec:appx_verification_frameworks}
\begin{table}[ht]
\centering
\small
\renewcommand{\arraystretch}{1.1}
\caption{\textbf{Verification failure mode vulnerabilities across frameworks.} \checkmark = vulnerable, -- = resilient, \checkmark/-- = partial. \textit{Frameworks:} TRL (Transformers RL)~\citep{vonWerra2024TRL}, VERL~\citep{VERL}, LM-Eval (hendrycks/minerva)~\citep{eval-harness}, LightEval~\citep{lighteval}, SCORE (LM-Eval SCORE math)~\citep{eval-harness}, evalchemy (ZeroEval)~\citep{Evalchemy}, HMMT (evalchemy matharena)~\citep{Evalchemy}, Math-V (Math-Verify)~\citep{Kydlicek2024MathVerify}.
\textit{Methods:} First\# = first number in string; Right = rightmost priority; Ltd. = limited support ($\leq$5 expressions).
\textit{F1--F7:} See Table~\ref{tab:failure-modes} for detailed descriptions.}\label{tab:all-frameworks-comparison}
\begin{tabular}{|l|c|c|c|c|c|c|c|c|}
\hline
\textbf{Failure Mode} & \textbf{TRL} & \textbf{VERL} & \textbf{LightEval} & \textbf{LM-Eval} & \textbf{SCORE} & \textbf{evalchemy} & \textbf{HMMT} & \textbf{Math-V} \\
\hline
F1: Multiple solutions (OR) & \checkmark & \checkmark & \checkmark/-- & \checkmark & \checkmark & \checkmark & \checkmark/-- & \checkmark/-- \\
\hline
F2: Late correct            & \checkmark & \checkmark/-- & \checkmark/-- & \checkmark/-- & -- & \checkmark & -- & \checkmark/-- \\
\hline
F3: Early correct           & -- & \checkmark & \checkmark & \checkmark & -- & -- & \checkmark & \checkmark \\
\hline
F4: Answer corrections      & \checkmark & -- & \checkmark & -- & -- & \checkmark & -- & \checkmark \\
\hline
F5: Unordered sets          & \checkmark & \checkmark & \checkmark & \checkmark & \checkmark & \checkmark & -- & \checkmark \\
\hline
F6: Missing anchors         & \checkmark & -- & -- & -- & -- & -- & \checkmark & -- \\
\hline
F7: MCQ partial match       & \checkmark & \checkmark & \checkmark & \checkmark & \checkmark & \checkmark & \checkmark & \checkmark \\
\hline
\textbf{Method}             & First & Last & Right & Last & All & First\# & Last/All & Right \\
\hline
\textbf{Anchors}            & Yes & No & No & No & No & No & No & Yes \\
\hline
\textbf{Multi-Answer}       & No & No & Ltd. & No & Partial & No & Yes & Ltd. \\
\hline
\end{tabular}
\end{table}

\subsection{Inference Parameters for Evaluation}\label{sec:appx_inference}
All models were run with nucleus sampling at a \texttt{top\_p} of 0.95. Other parameters were set according to model-specific recommendations:
\begin{itemize}
    \item \textbf{OLMo Models:} Temperature of 1.0, with \texttt{max\_tokens} set to 2048 (v1) and 4096 (v2).
    \item \textbf{Qwen2.5-Math Models:} Temperature of 0.6 and \texttt{max\_tokens} of 4096.
    \item \textbf{All Other Models:} Temperature of 0.6 and \texttt{max\_tokens} of 32,768.
\end{itemize}

\subsection{MATH-B details}
\begin{figure}[h!]
    \centering
    \includegraphics[width=1.0\linewidth]{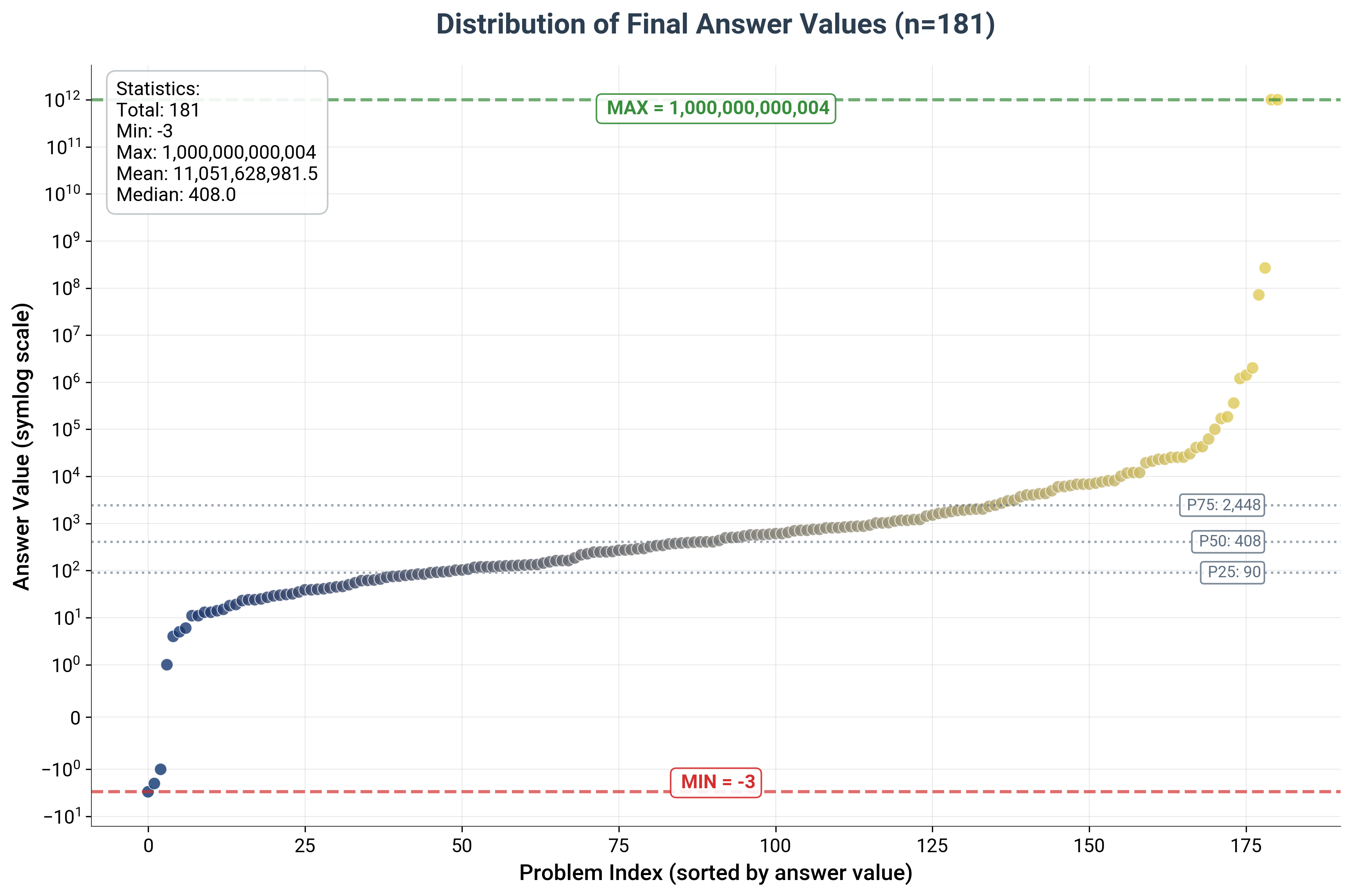}
    \caption{\textbf{Distribution of ground truth (final-answers) in MATH-B-U}. We use the log-scale for better readability.}
    \label{fig:final_answer}
\end{figure}

In \Cref{fig:final_answer}, we show the distribution of final answers for \textsc{MATH-B-U}, which spans a broad range of integer values.

\subsubsection{Looking at some samples from MATH-B}
In \Cref{tab:questions-intersection,tab:questions-intersection-hard}, we list ten randomly sampled \textsc{MATH-B-I} questions—the base split and the full set, respectively.

\begin{table}[ht]
\centering
\small
\renewcommand{\arraystretch}{1.2}
\caption{\textbf{Models' number of unsolved questions at \texttt{pass@1024}}. 
Models are grouped into Base vs. Supplementary. 
“Intersection (base)” and “Intersection (all)” indicate the overlap of unsolved problems across models.}
\label{tab:unsolved-pass1024-all}
\begin{tabular}{|l|c|c|c|}
\hline
\textbf{Model} & \textbf{\# Unsolved} & \textbf{Type} & \textbf{Notes} \\
\hline
Qwen2.5-1.5B & 145 & Base &  \\
Qwen2.5-7B & 120 & Base &  \\
Qwen2.5-Math-1.5B & 115 & Base &  \\
Qwen2.5-Math-7B & 101 & Base &  \\
Qwen3-4B-Base & 112 & Base &  \\
Qwen3-8B-Base & 116 & Base &  \\
DeepSeek-R1-Qwen2.5-1.5B & 115 & Base & Distill \\
DeepSeek-R1-Qwen2.5-7B & 99  & Base & Distill \\
OLMo-7B & 158 & Base &  \\
OLMo-2-7B & 132 & Base &  \\
Llama-3.1-8B & 151 & Base &  \\
\hline
Qwen2.5-1.5B-Instruct & 124 & Supplementary &  \\
Qwen2.5-7B-Instruct & 131 & Supplementary &  \\
Qwen2.5-Math-1.5B-Instruct & 139 & Supplementary &  \\
Qwen2.5-Math-7B-Instruct & 117 & Supplementary &  \\
Qwen3-4B & 67 & Supplementary &  \\
Qwen3-8B & 52 & Supplementary &  \\
DeepScaler-1.5B & 142 & Supplementary &  \\
Nemotron-1.5B-v1 & 137 & Supplementary &  \\
Nemotron-1.5B-v2 & 142 & Supplementary &  \\
Skywork-OR1-7B & 103 & Supplementary &  \\
\hline
\textbf{Intersection (base)} & 41 & -- & Shared failures across all base models \\
\textbf{Intersection (all)} & 13 & -- & Shared failures across all models \\
\hline
\end{tabular}
\end{table}

\begin{table}[ht]
\centering
\footnotesize
\renewcommand{\arraystretch}{1.2}
\caption{\textbf{Question texts (verbatim) from MATH-B-I (base)}. Each question includes the sentence ``Let's think step by step and output the final answer within \texttt{\textbackslash boxed\{\}}.'' appended.}
\label{tab:questions-intersection}
\begin{tabular}{|c|p{0.65\textwidth}|c|c|}
\hline
\textbf{ID} & \textbf{Question (wrapped to fit)} & \textbf{Difficulty} & \textbf{Final Answer} \\
\hline
1 & Fisica and Ritmo discovered a piece of Notalium shaped like a rectangular box, and wanted to find its volume. To do so, Fisica measured its three dimensions using a ruler with infinite precision, multiplied the results and rounded the product to the nearest cubic centimeter, getting a result of 2017 cubic centimeters. Ritmo, on the other hand, measured each dimension to the nearest centimeter and multiplied the rounded measurements, getting a result of $V$ cubic centimeters. Find the positive difference between the least and greatest possible positive values for $V$. & 4.5 & \multicolumn{1}{c|}{$\boxed{7174}$} \\
\hline
2 & Let $\mathbb{N}$ denote the natural numbers. Compute the number of functions $f:\mathbb{N}\rightarrow \{0,1,\dots,16\}$ such that $f(x+17)=f(x)$ and $f(x^2)\equiv f(x)^2+15 \pmod{17}$ for all integers $x\ge 1$. & 6.0 & \multicolumn{1}{c|}{$\boxed{12066}$} \\
\hline
3 & Each of the distinct letters in the following subtraction problem represents a different digit. Find the number represented by the word TEAM. PURPLE $-$ COMET $=$ TEAM. & 3.5 & \multicolumn{1}{c|}{$\boxed{6852}$} \\
\hline
4 & Find the sum of all positive integers $n$ such that there exists an integer $b$ with $|b|\neq 4$ such that the base $-4$ representation of $n$ is the same as the base $b$ representation of $n$. & 4.0 & \multicolumn{1}{c|}{$\boxed{1026}$} \\
\hline
5 & Let $S$ be a set of size $11$. A random $12$-tuple $(s_1,\dots,s_{12})$ of elements of $S$ is chosen uniformly at random. Moreover, let $\pi:S\to S$ be a permutation of $S$ chosen uniformly at random. The probability that $s_{i+1}\ne \pi(s_i)$ for all $1\le i \le 12$ (where $s_{13}=s_1$) can be written as $\tfrac{a}{b}$ where $a$ and $b$ are relatively prime positive integers. Compute $a$. & 6.0 & \multicolumn{1}{c|}{$\boxed{1000000000004}$} \\
\hline
6 & Let $\mathbb{N}$ denote the natural numbers. Compute the number of functions $f:\mathbb{N}\to\{0,1,\dots,16\}$ such that $f(x+17)=f(x)$ and $f(x^2)\equiv f(x)^2+15 \pmod{17}$ for all integers $x\ge 1$. & 4.5 & \multicolumn{1}{c|}{$\boxed{12066}$} \\
\hline
7 & A unicorn is tethered by a 20-foot rope to the base of a cylindrical tower of radius 8 feet. The rope is attached to the tower at ground level and to the unicorn at a height of 4 feet. The rope is taut, its end is 4 feet from the nearest point on the tower, and the length of rope touching the tower is $\tfrac{a-\sqrt{b}}{c}$ feet, where $a,b,c$ are positive integers and $c$ is prime. Find $a+b+c$. & 3.5 & \multicolumn{1}{c|}{$\boxed{813}$} \\
\hline
8 & For which maximal $N$ does there exist an $N$-digit number such that among any sequence of consecutive decimal digits some digit is present only once? & 6.0 & \multicolumn{1}{c|}{$\boxed{1023}$} \\
\hline
9 & Let $P=\{(x,y)\mid 0\le x,y\le 25,\ x,y\in\mathbb{Z}\}$. Let $T$ be the set of triangles formed by picking three distinct points in $P$ (rotations, reflections, and translations count as distinct). Compute the number of triangles in $T$ with area larger than 300. & 1.5 & \multicolumn{1}{c|}{$\boxed{436}$} \\
\hline
10 & Katie has a chocolate bar that is a $5$-by-$5$ grid of square pieces, but she only wants to eat the center piece. She repeatedly (i) chooses a gridline and splits the bar, (ii) discards the part not containing the center, (iii) repeats until only the center piece remains. Compute the number of possible sequences of operations. & 3.0 & \multicolumn{1}{c|}{$\boxed{6384}$} \\
\hline
\end{tabular}
\end{table}

\begin{table}[ht]
\centering
\footnotesize
\renewcommand{\arraystretch}{1.2}
\caption{\textbf{Question texts (verbatim) from MATH-B-I (all)} set of all considered models in \Cref{sec:final_pass1k_filtering}. Each question includes the sentence ``Let's think step by step and output the final answer within \texttt{\textbackslash boxed\{\}}.'' appended.}
\label{tab:questions-intersection-hard}
\begin{tabular}{|c|p{0.65\textwidth}|c|c|}
\hline
\textbf{ID} & \textbf{Question (wrapped to fit)} & \textbf{Difficulty} & \textbf{Final Answer} \\
\hline
1 & Bob is writing a sequence of letters of the alphabet, each of which can be either uppercase or lowercase, according to the following two rules: If he had just written an uppercase letter, he can either write the same letter in lowercase after it, or the next letter of the alphabet in uppercase. If he had just written a lowercase letter, he can either write the same letter in uppercase after it, or the preceding letter of the alphabet in lowercase. For instance, one such sequence is $a A a A B C D d c b B C$. How many sequences of 32 letters can he write that start at (lowercase) $a$ and end at (lowercase) $z$? & 1.5 & \multicolumn{1}{c|}{$\boxed{376}$} \\
\hline
2 & Let $V=\{1,\dots,8\}$. How many permutations $\sigma:V\to V$ are automorphisms of some tree? (A graph consists of some set of vertices and some edges between pairs of distinct vertices. It is connected if every two vertices in it are connected by some path of one or more edges. A tree $G$ on $V$ is a connected graph with vertex set $V$ and exactly $|V|-1$ edges, and an automorphism of $G$ is a permutation $\sigma:V\to V$ such that vertices $i,j\in V$ are connected by an edge iff $\sigma(i)$ and $\sigma(j)$ are.) & 5.5 & \multicolumn{1}{c|}{$\boxed{30212}$} \\
\hline
3 & Find the sum of all positive integers $n$ such that there exists an integer $b$ with $|b|\neq 4$ such that the base $-4$ representation of $n$ is the same as the base $b$ representation of $n$. & 4.0 & \multicolumn{1}{c|}{$\boxed{1026}$} \\
\hline
4 & A unicorn is tethered by a 20-foot silver rope to the base of a magician's cylindrical tower whose radius is 8 feet. The rope is attached to the tower at ground level and to the unicorn at a height of 4 feet. The rope is taut, the end of the rope is 4 feet from the nearest point on the tower, and the length of the rope touching the tower is $\tfrac{a-\sqrt{b}}{c}$ feet, where $a,b,c$ are positive integers and $c$ is prime. Find $a+b+c$. & 3.5 & \multicolumn{1}{c|}{$\boxed{813}$} \\
\hline
5 & A 16x16 square sheet of paper is folded once in half horizontally and once in half vertically to make an 8x8 square. This square is again folded in half twice to make a 4x4 square. This square is folded in half twice to make a 2x2 square. This square is folded in half twice to make a 1x1 square. Finally, a scissor is used to make cuts through both diagonals of all the layers of the 1x1 square. How many pieces of paper result? & 6.0 & \multicolumn{1}{c|}{$\boxed{544}$} \\
\hline
6 & Matt writes a permutation of $\{1,2,3,\dots,10\}$ across his paper with the leftmost number equal to 1 and the rightmost equal to 10. Exactly one interior number (not including 1 or 10) is less than both its immediate left and right neighbors. How many such permutations are there? & 6.0 & \multicolumn{1}{c|}{$\boxed{1636}$} \\
\hline
7 & Let $\mathbb{N}$ denote the natural numbers. Compute the number of functions $f:\mathbb{N}\rightarrow\{0,1,\ldots,16\}$ such that $f(x+17)=f(x)$ and $f(x^2)\equiv f(x)^2+15 \pmod{17}$ for all integers $x\ge 1$. & 4.5 & \multicolumn{1}{c|}{$\boxed{12066}$} \\
\hline
8 & Sir Alex plays the following game on a row of 9 cells. Initially all cells are empty. In each move he either (1) chooses any number of the form $2^j$ (nonnegative $j$) and puts it into an empty cell, or (2) chooses two cells with the same number $2^j$, replaces the number in one cell with $2^{j+1}$ and erases the number in the other cell. At the end, one cell contains $2^n$ and the others are empty. Determine the maximum number of moves possible in terms of $n$. Provide the value when $n=10$. & 5.0 & \multicolumn{1}{c|}{$\boxed{2025}$} \\
\hline
9 & Alice and Bob play on a board of one row of 2022 consecutive squares. They take turns placing domino tiles that cover two adjacent squares; Alice goes first. A tile must not cover a square already covered. The game ends when no tile can be placed. Alice wants to maximize the number of uncovered squares at the end; Bob wants to minimize it. What is the greatest number of uncovered squares Alice can ensure, no matter how Bob plays? & 4.0 & \multicolumn{1}{c|}{$\boxed{290}$} \\
\hline
10 & (Caos) A cao [sic] has 6 legs, 3 on each side. A walking pattern is an ordered sequence of raising and lowering each of the legs exactly once (total 12 actions), starting and ending with all legs on the ground. The pattern is safe if at any point he has at least 3 legs on the ground and not all three legs are on the same side. Estimate $N$, the number of safe patterns. & 4.0 & \multicolumn{1}{c|}{$\boxed{1416528}$} \\
\hline
\end{tabular}
\end{table}

\subsection{Analysis of \texttt{pass@k} Performance Scaling}

To further justify our evaluation methodology, we analyze the performance scaling of all 21 models from \Cref{tab:unsolved-pass1024-all} on the \textsc{MATH-B} Union set.

\begin{figure}[h!]
    \centering
    \includegraphics[width=1.0\linewidth]{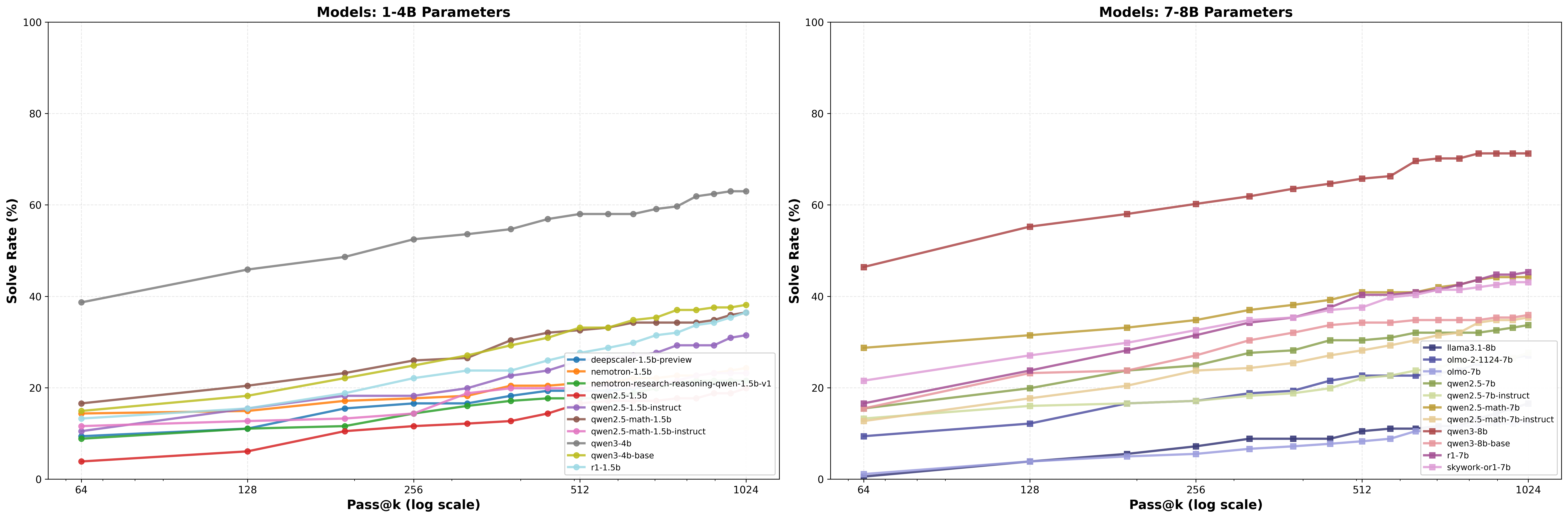}
    \caption{\textbf{Evolution of \texttt{pass@k} for all models on MATH-B-U}.}
    \label{fig:all_models_passk_curves}
\end{figure}

In \Cref{fig:all_models_passk_curves}, we plot the complete \texttt{pass@k} evolution as the sampling budget $k$ increases up to 1024. The performance curves for nearly all models exhibit a characteristic log-linear growth, indicating that improvement is consistent but not linear with computational effort. While this trend suggests continued gains with more sampling, it also shows initial signs of plateauing at higher values of $k$.

\begin{figure}[h!]
    \centering
    \includegraphics[width=1.0\linewidth]{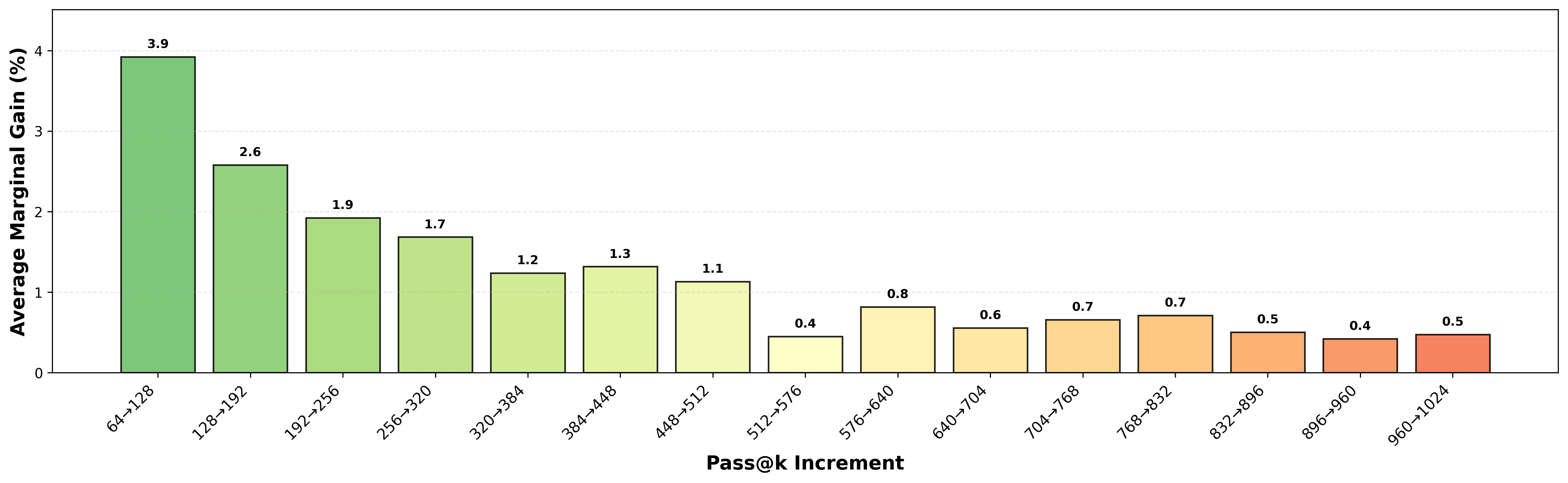}
    \caption{\textbf{Average gains in \texttt{pass@k} relative to the size of MATH-B-U.} Averaged over 21 models, the rate of solving new problems per 64-sample increment decreases as the total budget $k$ grows, demonstrating diminishing returns.}
    \label{fig:gains_per_k}
\end{figure}

To better quantify this observation, \Cref{fig:gains_per_k} visualizes the marginal gain in performance. Specifically, it plots the average increase in the \texttt{pass@k} rate for each successive 64-sample increment. This plot clearly illustrates the principle of diminishing returns: the most significant gains are concentrated at lower sampling budgets, and the rate of improvement slows considerably as the budget approaches 1024. Together, these figures provide strong empirical support for our choice of $k=1024$ as a practical and stable point for evaluation, beyond which brute-force sampling becomes an increasingly inefficient path to solving the remaining hard problems.

\subsection{LLM Usage}
The authors of this submission use IDEs with built-in LLM support, so LLMs have been used to help with menial coding tasks. Further, we used models like GPT-5, Gemini for rephrasing several paragraphs in this manuscript. In addition to that, we used GPT-5-mini and o4-mini-high to label the difficulty and topics of the benchmark we create (\Cref{sec:dataset_creation}).

\end{document}